\documentclass[11pt,a4paper]{article}
\usepackage[hyperref]{acl2021}
\usepackage{times}
\usepackage{latexsym}
\usepackage{graphicx}
\usepackage{multirow}
\usepackage{subcaption}
\usepackage{amsfonts}

\usepackage{microtype}

\aclfinalcopy 
\def\aclpaperid{2856} 

\title{{I}mproving Paraphrase Detection with the {A}dversarial {P}araphrasing {T}ask}

\author{Animesh Nighojkar, John Licato \\ Advancing Machine and Human Reasoning Lab \\
  Department of Computer Science and Engineering \\
  University of South Florida \\ Tampa, FL, USA \\
  \texttt{\{anighojkar,licato\}@usf.edu}}

\date{}

\begin{document}
\maketitle
\begin{abstract}
    If two sentences have the same meaning, it should follow that they are equivalent in their inferential properties, i.e., each sentence should textually entail the other. However, many paraphrase datasets currently in widespread use rely on a sense of paraphrase based on word overlap and syntax. Can we teach them instead to identify paraphrases in a way that draws on the inferential properties of the sentences, and is not over-reliant on lexical and syntactic similarities of a sentence pair? We apply the adversarial paradigm to this question, and introduce a new adversarial method of dataset creation for paraphrase identification: the Adversarial Paraphrasing Task (APT), which asks participants to generate semantically equivalent (in the sense of mutually implicative) but lexically and syntactically disparate paraphrases. These sentence pairs can then be used both to test paraphrase identification models (which get barely random accuracy) and then improve their performance. To accelerate dataset generation, we explore automation of APT using T5, and show that the resulting dataset also improves accuracy. We discuss implications for paraphrase detection and release our dataset in the hope of making paraphrase detection models better able to detect sentence-level meaning equivalence.
\end{abstract}

\section{Introduction}
\label{section-intro}

Although there are many definitions of `paraphrase' in the NLP literature, most maintain that two sentences that are paraphrases have the same meaning or contain the same information. \citet{pang-etal-2003-syntax} define paraphrasing as ``expressing the same information in multiple ways" and \citet{bannard2005paraphrasing} call paraphrases ``alternative ways of conveying the same information." \citet{ganitkevitch2013ppdb} write that ``paraphrases are differing textual realizations of the same meaning." A definition that seems to sufficiently encompass the others is given by \citet{bhagat-hovy-2013-squibs}: ``paraphrases are sentences or phrases that \textit{use different wording to convey the same meaning}." However, even that definition is somewhat imprecise, as it lacks clarity on what it assumes `meaning' means.

If paraphrasing is a property that can hold between sentence pairs,\footnote{In this paper we study paraphrase between sentences, and do not address the larger scope of how our work might extend to paraphrasing between arbitrarily large text sequences.} then it is reasonable to assume that sentences that are paraphrases must have equivalent meanings at the sentence level (rather than exclusively at the levels of individual word meanings or syntactic structures). Here a useful test is that recommended by inferential role semantics or inferentialism \cite{Boghossian1994, peregrin2006meaning}, which suggests that the meaning of a statement \textit{s} is grounded in its inferential properties: what one can infer from \textit{s} and from what \textit{s} can be inferred. 

Building on this concept from inferentialism, we assert that if two sentences have the same inferential properties, then they should also be mutually implicative. Mutual Implication (MI) is a binary relationship between two sentences that holds when each sentence textually entails the other (i.e., bidirectional entailment). MI is an attractive way of operationalizing the notion of two sentences having ``the same meaning,'' as it focuses on inferential relationships between sentences (properties of the sentences as wholes) instead of just syntactic or lexical similarities (properties of parts of the sentences). As such, we will assume in this paper that two sentences are paraphrases if and only if they are $MI$.\footnote{The notations used in this paper are listed in Table \ref{tbl:notations}.} 
In NLP, modeling inferential relationships between sentences is the goal of the textual entailment, or natural language inference (NLI) tasks \cite{bowman2015large}. 
We test MI using the version of RoBERTa$_{large}$ released by \citet{nie2020adversarial} trained on a combination of SNLI \cite{bowman2015large}, multiNLI \cite{williams2018broadcoverage}, FEVER-NLI \cite{nie2019combining}, and ANLI \cite{nie2020adversarial}.

Owing to expeditious progress in NLP research, performance of models on benchmark datasets is `plateauing' --- with near-human performance often achieved within a year or two of their release --- and newer versions, using a different approach, are constantly having to be created, for instance, GLUE \cite{wang2019glue} and SuperGLUE \cite{wang2020superglue}. The adversarial paradigm of dataset creation \cite{jia2017adversarial, jia-liang-2017-adversarial, bras2020adversarial, nie2020adversarial} has been widely used to address this `plateauing,' and the ideas presented in this paper draw inspiration from it. In the remainder of this paper, we apply the adversarial paradigm to the problem of paraphrase detection, and demonstrate the following \textbf{novel contributions}:

\begin{itemize}
    \item We use the adversarial paradigm to create a new benchmark examining whether paraphrase detection models are assessing the meaning equivalence of sentences rather than being over-reliant on word-level measures. We do this by collecting paraphrases that are \textit{MI} but are as lexically and syntactically disparate as possible (as measured by low BLEURT scores). We call this the Adversarial Paraphrasing Task (APT).
    \item We show that a SOTA language model trained on paraphrase datasets perform poorly on our benchmark. However, when further trained on our adversarially-generated datasets, their MCC scores improve by up to 0.307.
    \item We create an additional dataset by training a paraphrase generation model to perform our adversarial task, creating another large dataset that further improves the paraphrase detection models' performance.
    \item We propose a way to create a machine-generated adversarial dataset and discuss ways to ensure it does not suffer from the plateauing that other datasets suffer from.
\end{itemize}


\begin{table}
\centering
\resizebox{.9\linewidth}{!}{
\begin{tabular}{|c||c|}
\hline
MI & \begin{tabular}{@{}c@{}}Concept of mutual implication\\ / bidirectional textual entailment)\end{tabular} \\[4pt]
\hline
$MI$ & \begin{tabular}{@{}c@{}}Property of being mutually implicative, \\ as determined by our NLI model \end{tabular} \\[4pt]
\hline
APT & Adversarial Paraphrasing Task \\[4pt]
\hline
$APT$ & \begin{tabular}{@{}c@{}}Property of passing the adversarial \\paraphrase test (see \S \ref{section-apt}) \end{tabular} \\[4pt]
\hline
$AP_H$ & Human-generated APT dataset \\[4pt]
\hline
$AP_{T5}$ & \begin{tabular}{@{}c@{}}T5$_{base}$-generated APT dataset\\(Note that $AP_{T5} = AP^{M}_{T5} \cup AP^{Tw}_{T5}$)\end{tabular} \\[4pt]
\hline
$AP^M_{T5}$ & MSRP subset of $AP_{T5}$ \\[4pt]
\hline
$AP^{Tw}_{T5}$ & TwitterPPDB subset of $AP_{T5}$ \\[4pt]
\hline
\end{tabular}
}
\caption{Notations used in the paper.}
\label{tbl:notations}
\end{table}

\section{Related Work}
\label{sec-related}
Paraphrase detection (given two sentences, predict whether they are paraphrases) \cite{zhang2005paraphrase, fernando2008semantic, socher2011dynamic, jia-etal-2020-ask} is an important task in the field of NLP, finding downstream applications in machine translation \cite{callison-burch-etal-2006-improved, apidianaki2018automated, mayhew2020simultaneous}, text summarization, plagiarism detection \cite{hunt2019paraphraseplaigiarism}, question answering, and sentence simplification \cite{guo2018dynamic}. Paraphrases have proven to be a crucial part of NLP and language education, with research showing that paraphrasing helps improve reading comprehension skills \cite{lee2003paraphraserc, hagaman2008paraphraserc}. Question paraphrasing is an important step in knowledge-based question answering systems for matching questions asked by users with knowledge-based assertions \cite{fader2014, yin2015}.

Paraphrase generation (given a sentence, generate its paraphrase) \cite{Gupta_Agarwal_Singh_Rai_2018} is an area of research benefiting paraphrase detection as well. Lately, many paraphrasing datasets have been introduced to be used for training and testing ML models for both paraphrase detection and generation. 
MSRP \cite{dolan2005automatically} contains 5801 sentence pairs, each labeled with a binary human judgment of paraphrase, created using heuristic extraction techniques along with an SVM-based classifier. These pairs were annotated by humans, who found 67\% of them to be semantically equivalent.
The English portion of PPDB \cite{ganitkevitch2013ppdb} contains over 220M paraphrase pairs generated by meaning-preserving syntactic transformations. Paraphrase pairs in PPDB 2.0 \cite{pavlick2015ppdb} include fine-grained entailment relations, word embedding similarities, and style annotations.
TwitterPPDB \cite{lan-etal-2017-continuously} consists of 51,524 sentence pairs captured from Twitter by linking tweets through shared URLs. This approach's merit is its simplicity as it involves neither a classifier nor a human-in-the-loop to generate paraphrases. Humans annotate the pairs, giving them a similarity score ranging from 1 to 6.

ParaNMT \cite{wieting-gimpel-2018-paranmt} was created by using neural machine translation to translate the English side of a Czech-English parallel corpus (CzEng 1.6 \cite{bojar2016czeng}), generating more than 50M English-English paraphrases. However, ParaNMT's use of machine translation models that are a few years old harms its utility \cite{Nighojkar_Licato_2021}, considering the rapid improvement in machine translation in the past few years. To rectify this, we use the google-translate library to translate the Czech side of roughly 300k CzEng2.0 \cite{kocmi2020announcing} sentence pairs ourselves. We call this dataset \textit{ParaParaNMT} (PPNMT for short, where the extra \textit{para-} prefix reflects its similarity to, and conceptual derivation from, ParaNMT). 

Some work has been done in improving the quality of paraphrase detectors by training them on a dataset with more lexical and syntactic diversity. \citet{thompson2020paraphrase} propose a paraphrase generation algorithm that penalizes the production of n-grams present in the source sentence. Our approach to doing this is with the APT, but this is something worth exploring. \citet{sokolov2020neural} use a machine translation model to generate paraphrases much like ParaNMT. An interesting application of paraphrasing has been discussed by \citet{mayhew2020simultaneous} who, given a sentence in one language, generate a diverse set of correct translations (paraphrases) that humans are likely to produce. In comparison, our work is focused on generating adversarial paraphrases that are likely to deceive a paraphrase detector, and models trained on the adversarial datasets we produce can be applied to \citeauthor{mayhew2020simultaneous}'s work too. 

ANLI \cite{nie2020adversarial}, a dataset designed for Natural Language Inference (NLI) \cite{bowman2015large}, was collected via an adversarial human-and-model-in-the-loop procedure where humans are given the task of duping the model into making a wrong prediction. The model then tries to learn how not to make the same mistakes. AFLite \cite{bras2020adversarial} adversarially filters dataset biases making sure that the models are not learning those biases. They show that model performance on SNLI \cite{bowman2015large} drops from 92\% to 62\% when biases were filtered out. However, their approach is to filter the dataset, which reduces its size, making model training more difficult. Our present work tries instead to generate adversarial examples to \textit{increase} dataset size. Other examples of adversarial datasets in NLP include work done by \citet{jia2017adversarial, zellers2018swag, zellers2019hellaswag}. Perhaps the closest to our work is PAWS \cite{zhang2019paws}, short for Paraphrase Adversaries from Word Scrambling. The idea behind PAWS is to create a dataset that has a high lexical overlap between sentence pairs without them being `paraphrases.' It has 108k paraphrase and non-paraphrase pairs with high lexical overlap pairs generated by controlled word swapping and back-translation, and human raters have judged whether or not they are paraphrases. Including PAWS in the training data has shown the state-of-the-art models' performance to jump from 40\% to 85\% on PAWS's test split. In comparison to the present work, PAWS does not explicitly incorporate inferential properties, and we seek paraphrases \textit{minimizing} lexical overlap.
\section{Adversarial Paraphrasing Task (APT)}
\label{section-apt}

\begin{figure*}[t]
    \centering
    \includegraphics[width=.85\textwidth]{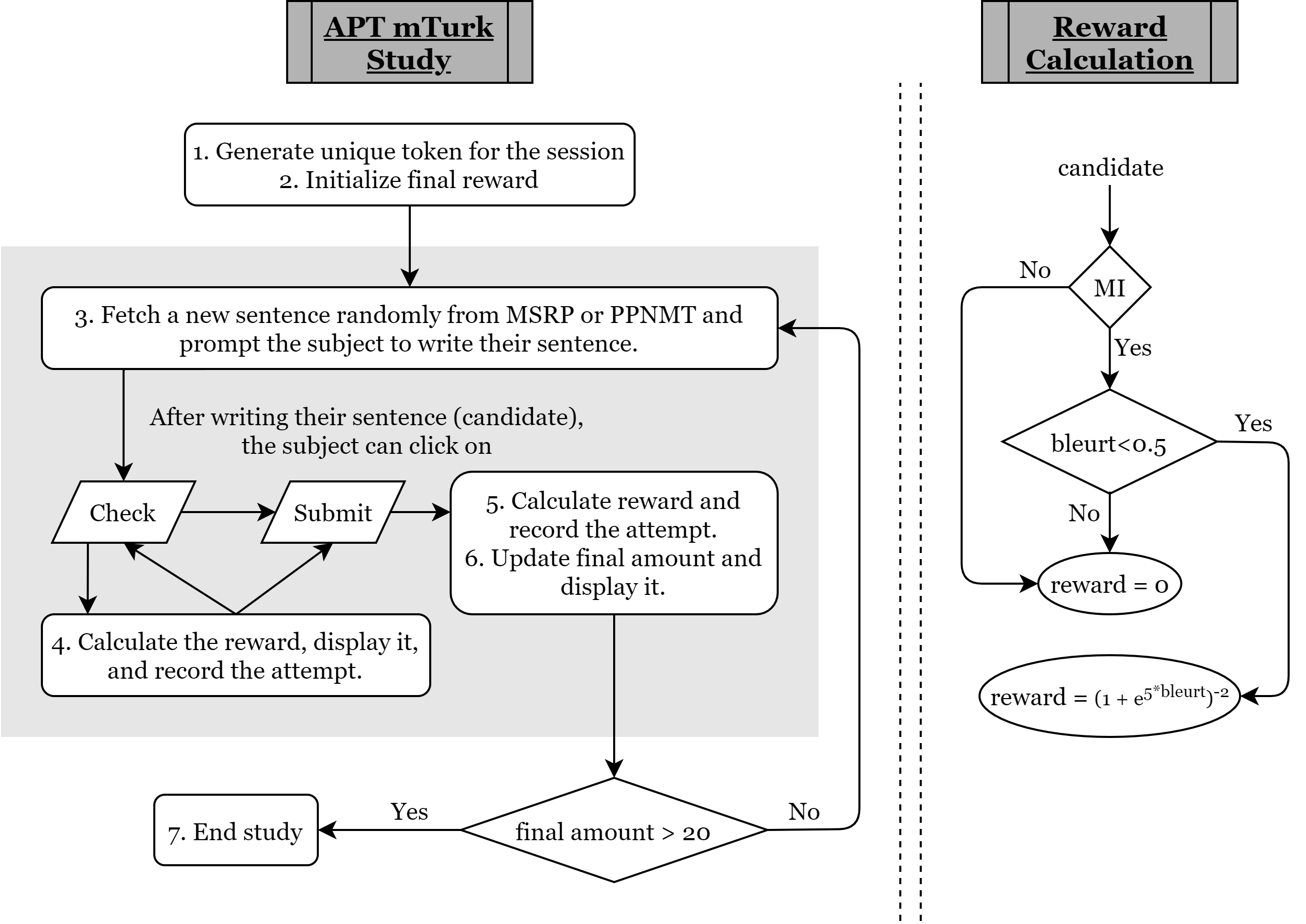}
    \caption{The mTurk study and the reward calculation. We automatically end the study when a subject earns a total of \$20 to ensure variation amongst subjects.}
    \label{fig:mturk}
\end{figure*}

Semantic Textual Similarity (STS) measures the degree of semantic similarity between two sentences.
Popular approaches to calculating STS include BLEU \cite{papineni-etal-2002-bleu}, BertScore \cite{zhang2020bertscore}, and BLEURT \cite{sellam2020bleurt}. BLEURT is a text generation metric building on BERT's \cite{devlin2019bert} contextual word representations. BLEURT is \textit{warmed-up} using synthetic sentence pairs and then fine-tuned on human ratings to generalize better than BERTScore \cite{zhang2020bertscore}. Given any two sentences, BLEURT assigns them a similarity score (usually between -2.2 to 1.1).
However, high STS scores do not necessarily predict whether two sentences have equivalent meanings. Consider the sentence pairs in Table \ref{tbl:examples}, highlighting cases where STS and paraphrase appear to misalign. The existence of such cases suggests a way to advance automated paraphrase detection: \textit{through an adversarial benchmark consisting of sentence pairs that have the same MI-based meaning, but have BLEURT scores that are as low as possible.} This is the motivation behind what we call the Adversarial Paraphrasing Task (APT), which has two components:
\begin{enumerate}
    \item \textit{Similarity of meaning}: Checked through MI (Section \ref{section-intro}). We assume if two sentences are $MI$ (Mutually Implicative), they are semantically equivalent and thus paraphrases. Note that MI is a binary relationship, so this APT component does not bring any quantitative variation but is more like a qualifier test for APT. All $APT$ sentence pairs are $MI$.
    \item \textit{Dissimilarity of structure}: Measured through BLEURT, which assigns each sentence pair a score quantifying how lexically and syntactically similar the two sentences are.
\end{enumerate}


\subsection{Manually Solving APT}

To test the effectiveness of APT in guiding the generation of mutually implicative but lexically and syntactically disparate paraphrases for a given sentence, we designed an Amazon Mechanical Turk (mTurk) study (Figure \ref{fig:mturk}). Given a starting sentence, we instructed participants to ``[w]rite a sentence that is the same in meaning as the given sentence but as structurally different as possible. Your sentence should be such that you can infer the given sentence from it AND vice-versa. It should be sufficiently different from the given sentence to get any reward for the submission. For example, a simple synonym substitution will most likely not work." The sentences given to the participants came from MSRP and PPNMT (Section \ref{section-intro}). Both of these datasets have pairs of sentences in each row, and we took only the first one to present to the participants. Neither of these datasets has duplicate sentences by design. Every time a sentence was selected, a random choice was made between MSRP and PPNMT, thus ensuring an even distribution of sentences from both datasets.

Each attempt was evaluated separately using Equation \ref{eq-dollar}, where $mi$ is 1 when the sentences are $MI$ and 0 otherwise:
\begin{equation}\label{eq-dollar}
    reward = \frac{mi}{(1 + e^{5*bleurt})^2}
\end{equation}
This formula was designed to ensure (1) the maximum reward per submission was \$1, and (2) no reward was granted for sentence pairs that are non-MI or have BLEURT $>0.5$. Participants were encouraged to frequently revise their sentences and click on a `Check' button which showed them the reward amount they would earn if they submitted this sentence. Once the `Check' button was clicked, the participant's reward was evaluated (see Figure \ref{fig:mturk}) and the sentence pair added to $AP_H$ (regardless of whether it was $APT$). If `Submit' was clicked, their attempt was rewarded based on Equation \ref{eq-dollar}.

\begin{table*}
\centering
\resizebox{.8\textwidth}{!}{
\begin{tabular}{|c||c|c|c|c||c|c|c|c|}
\hline
Dataset &
\begin{tabular}{@{}c@{}}Total \\ attempts\end{tabular} &    \begin{tabular}{@{}c@{}}$APT$ \\ attempts\end{tabular} &  \begin{tabular}{@{}c@{}}$MI$ \\ attempts\end{tabular} &   \begin{tabular}{@{}c@{}}non-$MI$ \\ attempts\end{tabular} &
\begin{tabular}{@{}c@{}}Unique \\ sentences\end{tabular} &  \begin{tabular}{@{}c@{}}$APT$ \\ uniques\end{tabular} &   \begin{tabular}{@{}c@{}}$MI$ \\ uniques\end{tabular} &    \begin{tabular}{@{}c@{}}non-$MI$ \\ uniques\end{tabular} \\
\hline
\hline
$AP_H$ &
5007 & \begin{tabular}{@{}c@{}}2659 \\ \textbf{53.10\%}\end{tabular} & \begin{tabular}{@{}c@{}}3232 \\ 64.55\%\end{tabular} & \begin{tabular}{@{}c@{}}1775 \\ 35.45\%\end{tabular} &
1631 & \begin{tabular}{@{}c@{}}1231 \\ \textbf{75.48\%}\end{tabular} & \begin{tabular}{@{}c@{}}1338 \\ 82.04\%\end{tabular} & \begin{tabular}{@{}c@{}}293 \\ 17.96\%\end{tabular} \\

$AP^M_{T5}$ &
62,986 & \begin{tabular}{@{}c@{}}3836 \\ \textbf{6.09\%}\end{tabular} & \begin{tabular}{@{}c@{}}37,511 \\ 59.55\%\end{tabular} & \begin{tabular}{@{}c@{}}25,475 \\ 40.44\%\end{tabular} &
4072 & \begin{tabular}{@{}c@{}}2288 \\ \textbf{56.19\%}\end{tabular} & \begin{tabular}{@{}c@{}}4045 \\ 99.34\%\end{tabular} & \begin{tabular}{@{}c@{}}3115 \\ 76.50\%\end{tabular} \\

$AP^{Tw}_{T5}$ &
75,011 & \begin{tabular}{@{}c@{}}6454 \\ \textbf{8.60\%}\end{tabular} & \begin{tabular}{@{}c@{}}17,074 \\ 22.76\%\end{tabular} & \begin{tabular}{@{}c@{}}57,937 \\ 77.24\%\end{tabular} &
4328 & \begin{tabular}{@{}c@{}}3670 \\ \textbf{84.80\%}\end{tabular} & \begin{tabular}{@{}c@{}}4131 \\ 95.45\%\end{tabular} & \begin{tabular}{@{}c@{}}4230 \\ 97.74\%\end{tabular} \\

\hline
\end{tabular}
}
\caption{Proportion of sentences generated by humans ($AP_H$) and T5$_{base}$ ($AP_{T5}$). ``Attempts" shows the number of attempts the participant made and ``Uniques" shows the number of source sentences from the dataset that the performer's attempts fall in that category on. For instance, 1631 unique sentences were presented to humans, who made a total of 5007 attempts to pass $APT$ and were able to do so for 2659 attempts which amounted to 1231 unique source sentences that could be paraphrased to pass $APT$.}
\label{tbl:proportion}
\end{table*}

The resulting dataset of sentence pairs, which we call $AP_H$ (Adversarial Paraphrase by Humans), consists of 5007 human-generated sentence pairs, both $MI$ and non-$MI$ (see Table \ref{tbl:proportion}). Humans were able to generate $APT$ paraphrases for $75.48\%$ of the sentences presented to them and only $53.1\%$ of attempts were $APT$, showing that the task is difficult even for humans. Note that `$MI$ attempts' and `$MI$ uniques' are supersets of `$APT$ attempts' and `$APT$ uniques,' respectively. 

\subsection{Automatically Solving APT}
\label{section-nap}

Since human studies can be time-consuming and costly, we trained a paraphrase generator to perform APT. We used T5$_{base}$ \cite{raffel2020exploring}, as it achieves SOTA on paraphrase generation \cite{niu2020unsupervised, bird2020chatbot, li2020agent} and trained it on TwitterPPDB (Section \ref{sec-related}). Our hypothesis was that if T5$_{base}$ is trained to maximize the APT reward (Equation \ref{eq-dollar}), its generated sentences will be more likely to be $APT$. We generated paraphrases for sentences in MSRP and those in TwitterPPDB itself, hoping that since T5$_{base}$ is trained on TwitterPPDB, it would generate better paraphrases ($MI$ with lower BLEURT) for sentences coming from there. The proportion of sentences generated by T5$_{base}$ is shown in Table \ref{tbl:proportion}. We call this dataset $AP_{T5}$, the generation of which involved two phases:

\noindent \textbf{Training:} To adapt T5$_{base}$ for APT, we implemented a custom loss function obtained from dividing the cross-entropy loss per batch by the total reward (again from Equation \ref{eq-dollar}) earned from the model's paraphrase generations for that batch, provided the model was able to reach a reward of at least 1. If not, the loss was equal to just the cross-entropy loss. We trained T5$_{base}$ on TwitterPPDB
for three epochs; each epoch took about 30 hours on one NVIDIA Tesla V100 GPU due to the CPU bound BLEURT component. More epochs \textit{may} help get better results, but our experiments showed that loss plateaus after three epochs.

\noindent \textbf{Generation:} Sampling, or randomly picking a next word according to its conditional probability distribution, introduces non-determinism in language generation. \citet{fan2018hierarchical} introduce top-k sampling, which filters $k$ most likely next words, and the probability mass is redistributed among only those $k$ words. Nucleus sampling (or top-p sampling) \cite{holtzman2020curious} reduces the options to the smallest possible set of words whose cumulative probability exceeds $p$, and the probability mass is redistributed among this set of words. Thus, the set of words changes dynamically according to the next word's probability distribution. We use a combination of top-k and top-p sampling with $k=120$ and $p=0.95$ in the interest of lexical and syntactic diversity in the paraphrases. For each sentence in the source dataset (MSRP\footnote{We use the official train split released by \citet{dolan2005automatically} containing 4076 sentence pairs.} and TwitterPPDB for $AP^M_{T5}$ and $AP^{Tw}_{T5}$ respectively), we perform five iterations, in each of which, we generate ten sentences. If at least one of these ten sentences passes $APT$, we continue to the next source sentence after recording all attempts and classifying them as $MI$ or non-$MI$. If no sentence in a maximum of 50 attempts passes $APT$, we record all attempts nonetheless, and move on to the next source sentence. For each increasing iteration for a particular source sentence, we increase $k$ by $20$, but we also reduce $p$ by $0.05$ to avoid vague guesses. Note the distribution of $MI$ and non-$MI$ in the source datasets does not matter because we use only the first sentence from the sentence pair.

\begin{figure*}[t]
    \centering
    \subfloat[Subfigure 1 list of figures text][$AP_H$]
        {
        \includegraphics[width=0.31\textwidth]{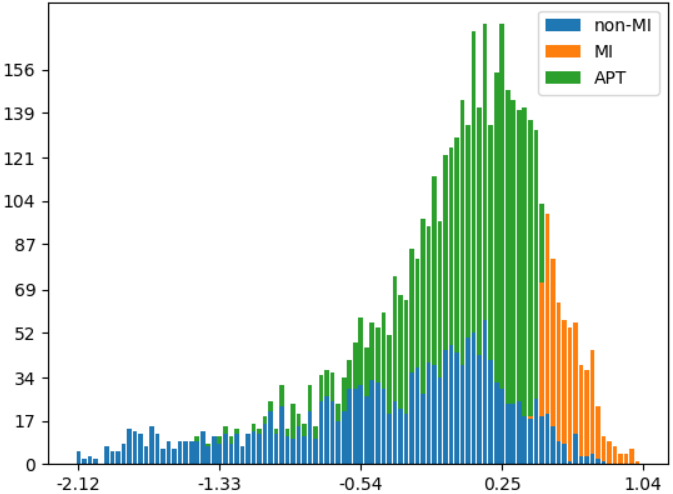}
        \label{fig:subfig1}
        }
    \subfloat[Subfigure 2 list of figures text][$AP^M_{T5}$]
        {
        \includegraphics[width=0.31\textwidth]{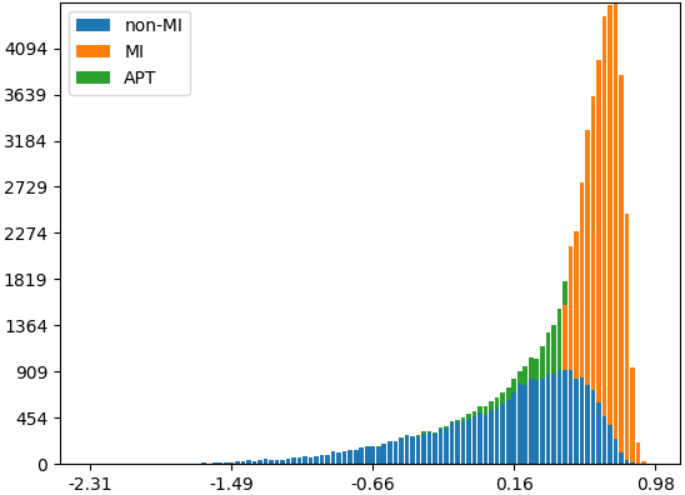}
        \label{fig:subfig2}
        }
    \subfloat[Subfigure 3 list of figures text][$AP^{Tw}_{T5}$]
        {
        \includegraphics[width=0.31\textwidth]{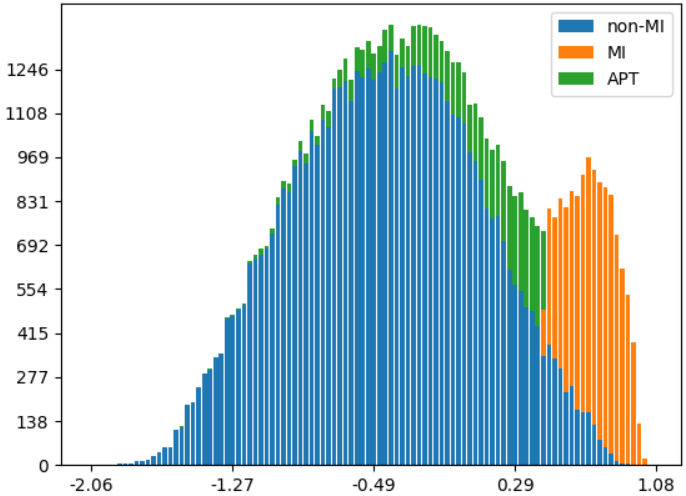}
        \label{fig:subfig3}
        }\\
    \caption{BLEURT distributions on adversarial datasets. All figures divide the range of observed scores into 100 bins. Note that $APT$ sentence pairs are also $MI$, whereas those labeled `MI' are not $APT$.}
    \label{fig:bleurt}
\end{figure*}

\vspace{-5pt}
\subsection{Dataset Properties}
T5$_{base}$ trained with our custom loss function generated $APT$-passing paraphrases for ($56.19\%$) of starting sentences. This is higher than we initially expected, considering how difficult APT proved to be for humans (Table \ref{tbl:proportion}). Noteworthy is that only $6.09\%$ of T5$_{base}$'s attempts were $APT$. This does not mean that the remaining $94\%$ of attempts can be discarded, since they amounted to the negative examples in the dataset.
Since we trained it on TwitterPPDB itself, we expected that T5$_{base}$ would generate better paraphrases, as measured by a higher chance of passing $APT$ on TwitterPPDB, than any other dataset we tested. This is supported by the data in Table \ref{tbl:proportion}, which shows that T5$_{base}$ was able to generate an $APT$ passing paraphrase for 84.8\% of the sentences in TwitterPPDB.

The composition of the three adversarial datasets can be found in Table \ref{tbl:proportion}. These metrics are useful to understand the capabilities of T5$_{base}$ as a paraphrase generator and the ``paraphrasability" of sentences in MSRP and TwitterPPDB. For instance, T5$_{base}$'s attempts on TwitterPPDB tend to be $MI$ much less frequently than those on MSRP and human's attempts on MSRP + PPNMT. This might be because in an attempt to generate syntactically dissimilar sentences, the T5$_{base}$ paraphraser also ended up generating many semantically dissimilar ones as well.

To visualize the syntactic and lexical disparity of paraphrases in the three adversarial datasets, we present their BLEURT distributions in Figure \ref{fig:bleurt}. As might be expected, the likelihood of a sentence pair being $MI$ increases as BLEURT score increases (recall that $APT$-passing sentence pairs are simply $MI$ pairs with BLEURT scores $<= 0.5$), but Figure \ref{fig:bleurt} shows that the shape of this increase is not straightforward, and differs among the three datasets. 


\begin{table*}
\centering
\resizebox{\textwidth}{!}{
\begin{tabular}{|c|c|c c|c|c|}
\hline
\textbf{No.} & \multirow{2}{*}{\begin{tabular}{@{}c@{}}\textbf{Source} \\ \textbf{Dataset}\end{tabular}} & \textbf{Source Sentence} & \textbf{Attempt} & \textbf{BLEURT} & \textbf{MI} \\
& & & & &\\
\hline
\hline
& & \multicolumn{2}{|c|}{$AP_H$} & & \\[4pt]
\cline{3-4}
1 & PPNMT &
So, can we please get out of here? & 
So is it okay if we please go? &
-0.064 & 1 \\[4pt]

2 & PPNMT &
You're crying. &
I did not cry &
-1.366 & 0 \\[4pt]

3 & PPNMT &
Treatment successful. &
The treatment was succesful. &
-0.871 & 1 \\[4pt]

4 & PPNMT &
Break a leg! &
Fracture a leg! &
0.408 & 1 \\[4pt]

5 & MSRP &
Two years later, the insurance coverage would begin. &
The insurance will start in two years &
0.281 & 1 \\[4pt]

6 & MSRP &
\multirow{2}{*}{\begin{tabular}{@{}c@{}}Evacuation went smoothly, although passengers \\ weren't told what was going on, Hunt said.\end{tabular}} &
Hunt told that Evacuation went smoothly. &
-0.298 & 0 \\
& & & & &\\[4pt]

\hline
\hline

& & \multicolumn{2}{|c|}{$AP_{T5}$} & & \\[4pt]
\cline{3-4}
7 & MSRP & 
Friday, Stanford (47-15) blanked the Gamecocks 8-0. &
Stanford (47-15) won 8-0 over the Gamecocks on Friday. &
0.206 & 1 \\[4pt]

8 & MSRP &
\multirow{2}{*}{\begin{tabular}{@{}c@{}}Revenue in the first quarter of the year dropped  \\ 15 percent from the same period a year earlier.\end{tabular}} &
\multirow{2}{*}{\begin{tabular}{@{}c@{}}Revenue declined 15 percent in the first quarter \\ of the year from the same period a year earlier.\end{tabular}} &
0.698 & 1 \\
& & & & &\\[4pt]

9 & MSRP &
\multirow{2}{*}{\begin{tabular}{@{}c@{}}A federal magistrate in Fort Lauderdale ordered \\ him held without bail.\end{tabular}} &
\multirow{2}{*}{\begin{tabular}{@{}c@{}}In Fort Lauderdale, Florida, a federal magistrate \\ ordered him held without bail.\end{tabular}} &
0.635 & 0 \\
& & & & &\\[4pt]

10 & TP &
\multirow{2}{*}{\begin{tabular}{@{}c@{}}16 innovations making a difference for poor \\ communities around the world.\end{tabular}} &
16 innovative ideas that tackle poverty around the world. &
0.317 & 1 \\
& & & & &\\[4pt]

11 & TP &
This is so past the bounds of normal or acceptable . &
This is so beyond the normal or acceptable boundaries. &
0.620 & 1 \\[4pt]

12 & TP &
\multirow{2}{*}{\begin{tabular}{@{}c@{}}The creator of Atari has launched a new VR company \\ called Modal VR.\end{tabular}} &
Atari creator is setting up a new VR company! &
0.106 & 0 \\
& & & & &\\[4pt]

\hline

\end{tabular}
}
\caption{Examples from adversarial datasets. The source dataset (TP short for TwitterPPDB) tells which dataset the sentence pair comes from (and whether it is in $AP^M_{T5}$ or $AP^{Tw}_{T5}$ for $AP_{T5}$). All datasets have $APT$ passing and failing $MI$ and non-$MI$ sentence pairs.}
\label{tbl:examples}
\end{table*}

As might be expected, humans are much more skilled at APT than T5$_{base}$, as shown by the fact that the paraphrases they generated have much lower mean BLEURT scores (Figure \ref{fig:bleurt}), and the ratio of $APT$ vs non-$APT$ sentences is much higher (Table \ref{tbl:proportion}). As we saw earlier, when T5$_{base}$ wrote paraphrases that were low on BLEURT, they tended to become non-$MI$ (e.g., line 12 in Table \ref{tbl:examples}). However, T5$_{base}$ did generate more $APT$-passing sentences with a lower BLEURT on Twitter-PPDB than on MSRP, which may be a result of overfitting T5$_{base}$ on TwitterPPDB. Furthermore, all three adversarial datasets have a distribution of $MI$ and non-$MI$ sentence pairs balanced enough to train a model to identify paraphrases.

Table 3 has examples from $AP_H$ and $AP_{T5}$ showing the merits and shortcomings of T5, BLEURT, and RoBERTa$_{large}$ (the MI detector used). Some observations from Table \ref{tbl:examples} include:
\begin{itemize}
    \item \textit{Lines 1 and 3:} BLEURT did not recognize the paraphrases, possibly due to the differences in words used. RoBERTa$_{large}$ however, gave the correct MI prediction (though it is worth noting that the sentences in line 1 are questions, rather than truth-apt propositions).
    \item \textit{Line 4:} RoBERTa$_{large}$ and BLEURT (to a large extent since it gave it a score of 0.4) did not recognize that the idiomatic phrase `break a leg' means `good luck' and not `fracture.'
    \item \textit{Lines 6 and 12:} There is a loss of information going from the first sentence to the second and BLEURT and MI both seem to have understood the difference between summarization and paraphrasing.
    \item \textit{Line 7:} T5 not only understood the scores but also managed to paraphrase it in such a way that was not syntactically and lexically similar, just as we wanted T5 to do when we fine-tuned it.
    \item \textit{Line 9:} T5$_{base}$ knows that Fort Lauderdale is in Florida but RoBERTa$_{large}$ does not.
\end{itemize}

\section{Experiments}
\label{sec-exp}

\begin{table}[t]
\centering
\resizebox{\linewidth}{!}{
\begin{tabular}{|c||c|cc|cc|}
\hline
\textbf{Dataset} & \textbf{Total} & \multicolumn{2}{|c|}{\textbf{$MI$}} & \multicolumn{2}{|c|}{\textbf{non-$MI$}} \\ 
\hline
\hline
$AP_H$-train & 3746 & 2433 & 64.95\% & 1313 & 35.05\% \\[4pt]

$AP_H$-test & 1261 & 799 & 63.36\% & 462 & 36.64\% \\[4pt]

MSRP-train & 4076 & 2753 & 67.54\% & 1323 & 32.46\% \\[4pt]

MSRP-test & 1725 & 1147 & 66.50\% & 578 & 33.50\% \\[4pt]
\hline
\end{tabular}
}
\caption{Distribution of $MI$ and non-$MI$ pairs.}
\label{tbl:mi-nmi}
\end{table}

\begin{table}[t]
\centering
\resizebox{0.8\linewidth}{!}{
\begin{tabular}{|c||c|c||c||c|c|}
\hline
\multirow{2}{*}{\textbf{Test Set}} & \multicolumn{2}{|c||}{\textbf{RoBERTa$_{base}$}} &
\multicolumn{2}{|c|}{\textbf{Random}} \\
\cline{2-5}
 & MCC & F1 & MCC & F1 \\
\hline
\hline
MSRP-train & 0.349 & 0.833 & 0 & 0.806 \\[4pt]

MSRP-test & 0.358 & 0.829 & 0 & 0.799 \\[4pt]

$AP_H$ & 0.222 & 0.746 & 0 & 0.784 \\[4pt]

$AP_H$-test & 0.218 & 0.743 & 0 & 0.777 \\[4pt]
\hline

\end{tabular}
}
\caption{Performance of RoBERTa$_{base}$ trained on just TwitterPPDB (no adversarial datasets) vs. random prediction.}
\label{tbl:roberta}
\end{table}




\begin{table}[t]
\centering
\resizebox{1.025\linewidth}{!}{
\begin{tabular}{|c|c||c|c||c|c|}
\hline
\multirow{2}{*}{\begin{tabular}{@{}c@{}}\textbf{Training Dataset} \\ TwitterPPDB + \end{tabular}} & \textbf{Size} & \multicolumn{2}{|c||}{\textbf{$AP_H$}} & \multicolumn{2}{|c|}{$AP_H$-test}  \\ 
\cline{3-6} & &  
MCC & F1 & MCC & F1 \\
\hline
\hline
$AP_H$-train & 46k & & & \textbf{0.440} & 0.809 \\[4pt]

$AP^M_{T5}$ & 106k & 0.410 & 0.725 & 0.369 & 0.705 \\[4pt]

$AP_H$-train + $AP^M_{T5}$ & 109k & & & \textbf{0.516} & 0.828 \\[4pt]

$AP^{Tw}_{T5}$ & 117k & 0.433 & 0.772 & 0.422 & 0.765 \\[4pt]

$AP_H$-train + $AP^{Tw}_{T5}$ & 121k & & & 0.488 & 0.812 \\[4pt]

$AP_{T5}$ & 180k & 0.461 & 0.731 & 0.437 & 0.716 \\[4pt]

$AP_H$-train + $AP_{T5}$ & 184k & & & \textbf{0.525} & 0.816 \\[4pt]
\hline
\end{tabular}
}
\caption{Performance of RoBERTa$_{base}$ trained on adversarial datasets. Size is the number of training examples in the dataset rounded to nearest 1000.}
\label{tbl:results}
\end{table}

To quantify our datasets' contributions, we designed experiment setups wherein we trained RoBERTa$_{base}$ \cite{liu2019roberta} for paraphrase detection on a combination of TwitterPPDB and our datasets as training data. RoBERTa was chosen for its generality, as it is a commonly used model in current NLP work and benchmarking, and currently achieves SOTA or near-SOTA results on a majority of NLP benchmark tasks \cite{wang2019glue, wang2020superglue, chen2021transformerbased}.

For each source sentence, multiple paraphrases may have been generated. Hence, to avoid data leakage, we created a train-test split on $AP_H$ such that all paraphrases generated using a given source sentence will be either in $AP_H$-train or in $AP_H$-test, but never in both. Note that $AP_H$ is not balanced as seen in Table \ref{tbl:proportion}. Table \ref{tbl:mi-nmi} shows the distribution of $MI$ and non-$MI$ pairs in $AP_H$-train and $AP_H$-test and `$MI$ attempts' and `non-$MI$ attempts' columns of Table \ref{tbl:proportion} show the same for other adversarial datasets. The test sets used were $AP_H$ wherever $AP_H$-train was not a part of the training data and $AP_H$-test in every case.

\paragraph{Does RoBERTa$_{base}$ do well on $AP_H$?}

RoBERTa$_{base}$ was trained on each training dataset (90\% training data, 10\% validation data) for five epochs with a batch size of 32 with the training and validation data shuffled, and the trained model was tested on $AP_H$ and $AP_H$-test. The results of this are shown in Table \ref{tbl:results}. Note that since the number of $MI$ and non-$MI$ sentences in all the datasets is imbalanced, Matthew's Correlation Coefficient (MCC) is a more appropriate performance measure than accuracy \cite{Boughorbel2017}.

Our motivation behind creating an adversarial dataset was to improve the performance of paraphrase detectors by ensuring they recognize paraphrases with low lexical overlap. To demonstrate the extent of their inability to do so, we first compare the performance of RoBERTa$_{base}$ trained only on TwitterPPDB on specific datasets as shown Table \ref{tbl:roberta}. Although the model performs slightly well on MSRP, it does barely better than a random prediction on $AP_H$, thus showing that identifying adversarial paraphrases created using APT is non-trivial for paraphrase identifiers.

\paragraph{Do human-generated adversarial paraphrases improve paraphrase detection?}
We introduce $AP_H$-train to the training dataset along with TwitterPPDB. This improves the MCC by 0.222 even though $AP_H$-train constituted just 8.15\% of the entire training dataset, the rest of which was TwitterPPDB (Table \ref{tbl:results}). This shows the effectiveness of human-generated paraphrases, as is especially impressive given the size of $AP_H$-train compared to TwitterPPDB.

\paragraph{Do machine-generated adversarial paraphrases improve paraphrase detection?}
We set out to test the improvement brought by $AP_{T5}$, of which we have two versions.
Adding $AP^M_{T5}$ to the training set was not as effective as adding $AP_H$-train, increasing MCC by 0.188 on $AP_H$ and 0.151 on $AP_H$-test, thus showing us that T5$_{base}$, although was able to clear $APT$, lacked the quality which human paraphrases possessed. This might be explained by Figure \ref{fig:bleurt} --- since $AP^M_{T5}$ does not have many sentences with low BLEURT, we cannot expect a vast improvement in RoBERTa$_{base}$'s performance on sentences with BLEURT as low as in $AP_H$. 
    
Since we were not necessarily testing T5$_{base}$'s performance --- and we had trained T5$_{base}$ on TwitterPPDB --- we used the trained model to perform APT on TwitterPPDB itself. Adhering to expectations, training RoBERTa$_{base}$ (the paraphrase detector) with $AP^{Tw}_{T5}$ yielded higher MCCs. Note that none of the sentences are common between $AP^{Tw}_{T5}$ and $AP_H$ since $AP_H$ is built on MSRP and PPNMT and the fact that the model got this performance when trained on $AP^{Tw}_{T5}$ is a testimony to the quality and contribution of APT.

Combining these results, we can conclude that although machine-generated datasets like $AP_{T5}$ can help paraphrase detectors improve themselves, a smaller dataset of human-generated adversarial paraphrases improved performance more. Overall, however, the highest MCC (0.525 in Table \ref{tbl:results}) is obtained when TwitterPPDB is combined with all three adversarial datasets, suggesting that the two approaches nicely complement each other. 

\section{Discussions and Conclusions}
This paper introduced APT (Adversarial Paraphrasing Task), a task that uses the adversarial paradigm to generate paraphrases consisting of sentences with equivalent (sentence-level) meanings, but differing lexical (word-level) and syntactical similarity. We used APT to create a human-generated dataset / benchmark ($AP_H$) and two machine-generated datasets ($AP_{T5}^M$ and $AP_{T5}^{Tw}$). 
Our goal was to effectively augment how paraphrase detectors are trained, in order to make them less reliant on word-level similarity. In this respect, the present work succeeded: we showed that RoBERTa$_{base}$ trained on TwitterPPDB performed poorly on APT benchmarks, but this performance was increased significantly when further trained on either our human- or machine-generated datasets. The code used in this paper along with the dataset has been released in a publicly-available repository.\footnote{\url{https://github.com/Advancing-Machine-Human-Reasoning-Lab/apt}}

Paraphrase detection and generation have broad applicability, but most of their potential lies in areas in which they still have not been substantially applied. These areas range from healthcare (improving accessibility to medical communications or concepts by automatically generating simpler language), writing (changing the writing style of an article to match phrasing a reader is better able to understand), and education (simplifying the language of a scientific paper or educational lesson to make it easier for students to understand). Thus, future research into improving their performance can be very valuable. But approaches to paraphrase that treat it as no more than a matter of detecting word similarity overlap will not suffice for these applications. Rather, the meanings of sentences are properties of the sentences as a whole, and are inseparably tied to their inferential properties. Thus, our approaches to paraphrase detection and generation must follow suit.

The adversarial paradigm can be used to dive deeper into comparing how humans and SOTA language models understand sentence meaning, as we did with APT. Furthermore, automatic generation of adversarial datasets has much unrealized potential; e.g., different datasets, paraphrase generators, and training approaches can be used to generate future versions of $AP_{T5}$ in order to produce $APT$ passing sentence pairs with lower lexical and syntactic similarities (as measured not only by BLEURT, but also by future state-of-the-art STS metrics).
The idea of more efficient automated adversarial task performance is particularly exciting, as it points to a way language models can improve themselves while avoiding prohibitively expensive human participant fees. 

Finally, the most significant contribution of this paper, APT, presents a dataset creation method for paraphrases that will not saturate because as the models get better at identifying paraphrases, we will improve paraphrase generation. As models get better at generating paraphrases, we can make APT harder (e.g., by reducing the BLEURT threshold of $<0.5$). One might think of this as students in a class who come up with new ways of copying their assignments from sources as plagiarism detectors improved. That brings us to one of the many applications of paraphrases: plagiarism generation and detection, which inherently is an adversarial activity. Until plagiarism detectors are trained on adversarial datasets themselves, we cannot expect them to capture human levels of adversarial paraphrasing.

\section*{Acknowledgements}
This material is based upon work supported by the Air Force Office of Scientific Research under award numbers FA9550-17-1-0191 and FA9550-18-1-0052. Any opinions, findings, and conclusions or recommendations expressed in this material are those of the authors and do not necessarily reflect the views of the United States Air Force. We would also like to thank Antonio Laverghetta Jr. and Jamshidbek Mirzakhalov for their helpful suggestions while writing this paper, and Gokul Shanth Raveendran and Manvi Nagdev for helping with the website used for the mTurk study.

\bibliographystyle{acl_natbib}
\bibliography{acl2021}


\end{document}